# VR-LENS: Super Learning-based Cybersickness Detection and Explainable AI-Guided Deployment in Virtual Reality

RIPAN KUMAR KUNDU, University of Missouri-Columbia, USA
OSAMA YAHIA ELSAID, University of Missouri-Columbia, USA
PRASAD CALYAM, University of Missouri-Columbia, USA
KHAZA ANUARUL HOQUE, University of Missouri-Columbia, USA

Virtual reality (VR) systems are known for their susceptibility to cybersickness, which can seriously hinder users' experience. Therefore, a plethora of recent research has proposed several automated methods based on machine learning (ML) and deep learning (DL) to detect cybersickness. However, these detection methods are perceived as computationally intensive and black-box methods. Thus, those techniques are neither trustworthy nor practical for deploying on standalone VR head-mounted displays (HMDs). This work presents an explainable artificial intelligence (XAI)-based framework *VR-LENS* for developing cybersickness detection ML models, explaining them, reducing their size, and deploying them in a Qualcomm Snapdragon 750G processor-based Samsung A52 device. Specifically, we first develop a novel super learning-based ensemble ML model for cybersickness detection. Next, we employ a post-hoc explanation method, such as SHapley Additive exPlanations (SHAP), Morris Sensitivity Analysis (MSA), Local Interpretable Model-Agnostic Explanations (LIME), and Partial Dependence Plot (PDP) to explain the expected results and identify the most dominant features. The super learner cybersickness model is then retrained using the identified dominant features. Our proposed method identified eye tracking, player position, and galvanic skin/heart rate response as the most dominant features for the integrated sensor, gameplay, and bio-physiological datasets. We also show that the proposed XAI-guided feature reduction significantly reduces the model training and inference time by 1.91X and 2.15X while maintaining baseline accuracy. For instance, using the integrated sensor dataset, our reduced super learner model outperforms the state-of-the-art works by classifying cybersickness into 4 classes (none, low, medium, and high) with an accuracy of 96% and regressing (FMS 1–10) with a Root Mean Square Error (RMSE) of 0.03. Our proposed method can help researchers analyze, detect, and mitigate cybersickness in real time and deploy the super learner-based cybersickness detection model in standalone VR headsets.

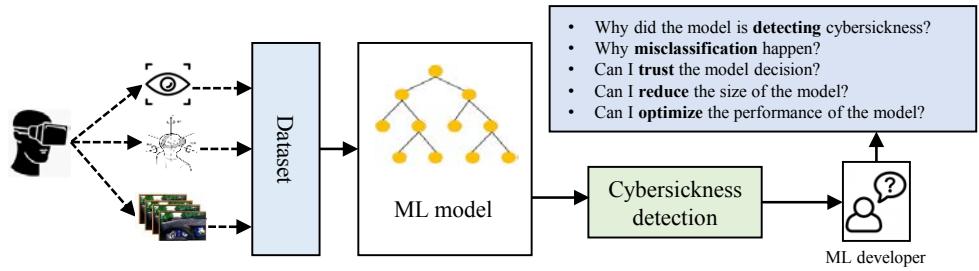

Fig. 1. Key questions in ML-based cybersickness (CS) detection.

## 1 INTRODUCTION

Virtual reality (VR) is being applied in various fields containing national defense [2], education [59], health care [27], public safety [53], and others [7, 19, 70, 76]. Specifically, in the current COVID-19 pandemic situation, VR offers a tremendous prospect for remote learning [63], as realistic collaboration workspaces [88, 98], and also as coping strategies for mental wellness for adults [19] as it can provide a sense of human presence. However, VR users often experience VR sickness or cybersickness, which hinders their immersive experience. Thus cybersickness has emerged as an important obstacle [59, 76, 87] to the wider acceptability of VR. Cybersickness can be defined as a set of unpleasant symptoms such as eyestrain, headache, nausea, disorientation, or even vomiting [22, 50, 70, 87]. One of the popular techniques to detect cybersickness is to use post-immersive subjective questionnaires such as Simulator Sickness Questionnaire (SSQ) and the VR Sickness Questionnaire (VRSQ) [84]. In contrast, the Fast Motion Sickness Scale (FMS) [43] can assess cybersickness severity during immersion. However, FMS relies on user feedback, i.e., needs human intervention during the VR immersion [62]. To overcome these limitations, deep learning (DL) and machine learning (ML) has recently become popular for cybersickness detection [1, 31, 32, 38, 41, 50, 75].

State-of-the-art ML/DL models can detect cybersickness with good accuracy from physiological signals (e.g., Heart Rate (HR), Galvanic Skin Responses (GSR), Breathe Rate (BR), Electroencephalogram (EEG)) [19, 65, 75, 91, 100], integrated eye and head tracking sensors, [33], and stereoscopic video [33, 52, 67]. For instance, Islam et al. [33] proposed a cybersickness severity detection using a deep fusion network with an accuracy of 87.7% from users' eye tracking and head tracking data. Other researchers used electroencephalography (EEG) [75, 91, 100], stereoscopic video [33, 67] and bio-physiological signal [31, 75] data for detecting cybersickness with good accuracy. Despite the great prospect of ML/DL models in detecting cybersickness, these methods have several drawbacks shown (see Figure 1).

- Most of the state-of-the-art ML/DL models for VR cybersickness detection rely on black-box models; thus, they lack explainability. Adding explainability to these black-box models can significantly improve the model's trustworthiness and provide insight into why and how the ML/DL model arrived at a specific decision.
- VR sensors generate a tremendous amount of data, resulting in complex, large, and power-hungry models. This makes real-time cybersickness detection challenging in standalone head-mounted devices (HMDs). Adding the explanation to the ML/DL model can guide the engineers to optimize the model effectively by identifying the dominating features. Indeed, reducing the dimensionality of cybersickness ML/DL models can significantly improve their training time, inference time, and size without compromising accuracy.

To demonstrate the need for the proposed XAI approach, consider that a VR game developer may want to develop a cybersickness detection model for their game based on alpha/beta players who are using a resource-constrained standalone VR headset. Suppose the developer used a black box cybersickness detection model without explainability in the game. In that case, they could not easily identify which of the model's features (e.g., gaze origin from world space) contributed to cybersickness prediction. Hence, they would have used trial and error to reduce the model size to minimize resource usage on the already resource-constrained Meta Quest Pro.

It is worth mentioning that there is a significant research gap in applying explainable artificial intelligence (XAI) to explain cybersickness. Very recently, Kundu et al. [49] used inherently interpretable ML models for cybersickness detection and explanation using physiological[35] and gameplay datasets[73]. However, their work considered binary classifiers, which can only detect the presence or absence of cybersickness. Such simple models are ineffective for realistic VR applications as they cannot detect cybersickness severity. Additionally, inherently interpretable models are typically dependent on the data properties and thus suffer from the curse of dimensionality problem [83]. For instance, decision tree and logistic regression-based inherently interpretable models can suffer from their overfitting problem because of their dimensionality and nonlinearity, which may eventually lead to their poor performance in cybersickness classification [49, 69, 75]. Most prior works use feature selection or dimensionality reduction techniques, such as principal component analysis (PCA) [45, 47, 60, 82, 89] to address the high-dimensionality issue in ML/DL models for cybersickness detection. For instance, Lin et al. [55] applied ML models with PCA to extract the cybersickness-related features to predict the cybersickness level. Similarly, Kottaimalai et al. in [48] used PCA to reduce dimensions, complexity, and computational time for EEG signals to detect cybersickness. However, applying PCA results in loss of information [3]. Furthermore, Mawalid et al. in [61] used time-domain feature extraction methods to extract the EEG statistical features for classifying cybersickness. However, applying PCA results in loss of information [3]. This means applying PCA-based dimensionality reduction may result in losing important features essential for accurate cybersickness detection.

To address the above-mentioned challenges, we propose a novel methodology, **VR-LENS**–an XAI-based framework for cybersickness detection, explanation, model reduction, and model deployment. First, to demonstrate the applicability of our proposed method, we proposed a novel super learning-based ensemble ML model. Then, we employed post-hoc explanation methods, SHapley Additive exPlanations (SHAP), Morris Sensitivity Analysis (MSA), Local Interpretable Model-Agnostic Explanations (LIME), and Partial Dependence Plot (PDP) to explain the expected results and identify the most dominant features. Specifically, we first develop a novel super learning-based ensemble ML model for cybersickness detection. Then, we employ post-hoc explanation methods, namely SHapley Additive exPlanations (SHAP) [57], Morris Sensitivity Analysis (MSA) [64], Local Interpretable Model-Agnostic Explanations (LIME) [80], and Partial Dependence Plot (PDP) [24], to provide global and local explanations for analyzing, identifying, and ranking dominating features causing cybersickness. The identified dominating features are then used to retrain the super learner model, i.e., to train them with a reduced number of features. This results in a lightweight cybersickness detection model with a significantly reduced number of features. Finally, to show the effectiveness of our VR-LENS framework, we deployed the reduced super learner model in a Qualcomm Snapdragon 750G processor-based Samsung A52 device [30] since many state-of-art VR devices are built using Qualcomm Snapdragon processors. We show that our proposed deployed reduced super learner model results in faster training and significantly faster inference time in the deployed device with great accuracy outperforming the state-of-the-art works. For instance, our results show that using the integrated sensor dataset [33], the proposed super learner model with all features classifies the cybersickness severity into 4 classes (none, low, medium, and high) with an accuracy of 95% and regress (FMS 1–10) the ongoing cybersickness with a Root Mean Square Error (RMSE) value of 0.04. However, after the XAI-based reduction of the same super learner model and the same dataset, we can classify the cybersickness severity and predict the ongoing cybersickness with an accuracy of 96% and RMSE of 0.03, respectively, with only 1/3 of the features of the baseline model. Furthermore, our reduction approach results in a 1.91X improvement in training time and a 2.15X improvement in the inference time in the deployed embedded device. To the best of our knowledge, this is the first work that uses a super learner model and post-hoc explanation techniques for cybersickness detection, explanations, model reduction, and deployment. Therefore, we believe the proposed method can help future researchers understand, analyze, and design more straightforward, lightweight, trustworthy, and accurate cybersickness detection models suitable for real-time implementation in standalone HMDs.

## 2 RELATED WORKS

The term *cybersickness* refers to a group of symptoms, such as dizziness, nausea, etc., that are similar to motion sickness and can occur during or after an immersive experience [51]. The most popular theory to explain the reason behind cybersickness is the sensory conflict [50, 51]. This theory states that cybersickness occurs when the eyes sense motion, but the vestibular system does not. However, other theories, such as poison theory and postural instability theory, have also been used to explain the causes of cybersickness. In addition, factors such as age, gender, and prior VR experience of users can also impact the degree of cybersickness [15, 21, 29]. Other factors causing cybersickness are display, latency, flickering, lag, cybersecurity, etc., [25, 38, 50, 93, 94]. The state-of-the-art works in cybersickness detection can be broadly divided into two categories using subjective, objective, and advanced ML/DL methods. In addition, there also exists work in the area of dimensionality reduction of cybersickness models. In this section, we discuss these related works as follows.

Researchers have proposed several subjective measurements such as the Simulator Sickness Questionnaire (SSQ) [8–10, 16, 84, 96], the FMS [43], and the Motion Sickness Susceptibility Questionnaire (MSSQ) [42] to measure cybersickness. In contrast, several researchers have also proposed objective measurements (i.e., physiological signals) for cybersickness [31, 35, 73] detection. Previous research has shown that objective measurements (e.g., heart rate, gastric tachyarrhythmia, galvanic skin response, eye-blink rate, pupil diameter and electroencephalogram (EEG))delta, and beta wave signals) vary significantly when cybersickness occurs [14, 35, 56, 56, 77, 79]. For instance, they found that HR and EEG delta waves correlate positively with cybersickness, whereas EEG beta waves correlate negatively [56]. On the contrary, another study reported that GSR has a stronger positive correlation with cybersickness than other objective measurements that can detect cybersickness [35, 90].

Numerous ML and DL-based approaches have recently been proposed [1, 26, 31, 33, 35, 37, 44, 50, 68, 75, 78, 79, 95, 97, 99] for detecting cybersickness automatically from a variety of subjective measurements (FMSQ, MSSQ) data, objective measurements (bio-physiological signals) data, and integrated multimodal sensors measurements (eye-tracking, head-tracking, motion-flow, etc.) data in HMD. For instance, Seungjun et al. [1] proposed a machine–deep–ensemble learning method for classifying the cybersickness from bio-physiological data. In contrast, in [73], a symbolic ML-based approach is used to identify the levels of cybersickness. Moreover, Azadeh et al. [26] used Topological Data Analysis (TDA) based on support vector machine (SVMs) with Gaussian RBF kernel methods for predicting cybersickness from physiological and subjective measurements data. On the other hand, Padmanaban et al. [67] used the ML algorithm on hand-crafted features from the VR video data to predict cybersickness. Apart from the ML-based method, in recent years DL-based method has gained more attention from cybersickness researchers. For instance, the authors in [41] applied three DL/ML-based methods, namely Convolutional neural network (CNN), LSTM, and Support Vector Regression (SVR) for cybersickness prediction. In contrast, Lee et al. [52] used a 3D-CNN and a multimodal deep fusion network to detect cybersickness using optical flow, disparity, and saliency features from the VR video data. On the other hand, Jeong et al. [37] applied attention-based DL models for predicting cybersickness from integrated sensor data. Likewise, in [75], an LSTM-based attention network is used for detecting cybersickness from user bio-physiological signals. Consequently, Islam et al. [35] applied an LSTM-based network to classify the cybersickness severity from users' bio-physiological data (e.g., HR, GSR, etc.). In addition, a deep fusion approach was presented in [33] for classifying cybersickness severity from the multimodal integrated sensors data (e.g., eye-tracking, head-tracking, etc.). Although ML/DL-based methods have shown enormous success in cybersickness detection, there is a significant research gap in applying XAI to explain detected cybersickness models. Indeed, understanding why some samples are correctly vs. incorrectly labeled as cybersickness and which feature contributed to the cybersickness detection result is an important step toward applying the proper mitigation technique. Therefore, explicit explanations are required to ascertain which feature (e.g., eye-tracking, head-tracking, HR, galvanic skin response, etc.) is responsible for cybersickness detection. Moreover, applying XAI in cybersickness detection models can significantly improve the model's *trustworthiness* and provide insight into why and how the ML/DL model arrived at a specific decision. In our context, the concept of trustworthy refers to using XAI for cybersickness explanations through mechanisms such as global and local explanations and explainable layers, which make the ML/DL model transparent, understandable, and therefore, trustworthy to users [92]. Recently, Kundu et al. [49] used three inherently interpretable ML models, namely explainable boosting Machine (EBM), decision tree (DT), and logistic regression (LR), to detect and explain the cybersickness from a user's bio-physiological and subjective measurement data. However, their proposed approach is limited to binary classification (cybersickness vs. no cybersickness).. In contrast, our work considers a multi-class classification problem (none, low, medium, and high) for cybersickness detection. Another limitation in [49] is that their proposed EBM model is required higher training time [66], which is not feasible for real-time deployment. In contrast, our work proposes a lightweight, super learner model and uses a post-hoc explanation-based method such as SHAP, MSA, LIME, and PDP to explain black-box ML model cybersickness detection.

There exist several works in dimensionality reduction of cybersickness models. Many researchers used PCA-based methods to identify important features and reduce the model size [45, 47, 60, 82, 85, 89]. For example, Lin et al. [55] used PCA based method to extract the cybersickness-related features from EEG signals. After feature extraction, they utilized 3 ML/DL models, namely, linear regression (LR), SVM, and self-organizing neural fuzzy inference network (SOFIN), to predict the user's level of cybersickness. Similarly, the authors in [48] also used PCA to find the patterns from the EEG signals and neural networks (NN) to classify the cognitive tasks using the Colorado University EEG signal dataset. In contrast, Singla et al. in [85] used PCA to reduce the set of questions from the SSQ simulation. On the other hand, Mawalid et al. in [61] used time domain feature extraction based on the statistical features (e.g., mean, variation, standard deviation, number of peaks) and power percentage band to understand the cybersickness features and then applied K-Nearest Neighbor and Naive Bayes classifiers to classify cybersickness. However, PCA-based dimension reduction is not always trustworthy and has a few drawbacks. For instance, PCA maps high-dimensional data to low-dimensional space through projections, which often leads to the loss of information from the original data[40]. In contrast, We use an XAI-based approach for dimensionality reduction of the cybersickness detection model to avoid the loss of information and maintain *trustworthy*. Indeed, it is important to identify the key features inducing cybersickness in VR to develop effective mitigation methods. This can be achieved by using XAI techniques. However, to the best of our knowledge, XAI techniques for detecting and predicting cybersickness have not been explored yet, which motivates our work in the paper.

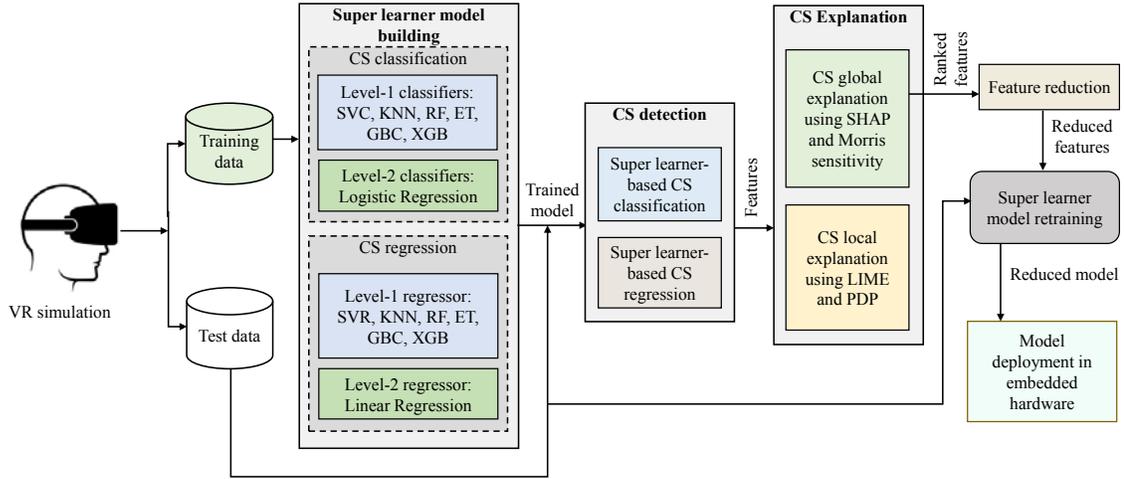

Fig. 2. Overview of VR-LENS for cybersickness (CS) detection, explanation, reduction, and deployment.

## 3 VR-LENS FRAMEWORK

An overview of the proposed VR-LENS framework for VR cybersickness detection, explanation, model reduction, and deployment is shown in Figure2. At first, the training data is used to train the super learner-based ensemble learning model for training for both classification and regression tasks (explained in detail in the next section). The cybersickness classification and regression training phase consist of two levels of the classifier. The level-1 classifier consists of multiple ML models as base models, and the level-2 classifier consists of a meta-classifier for the final classification. Next, the trained VR-LENS-based classification and regression models are used for classifying and regressing the cybersickness from the test dataset. The cybersickness regression aims at classifying the cybersickness levels with the trained super learner model. In contrast, the cybersickness regression predicts the next value of the ongoing cybersickness FMS score in the range of 0 to 10. The next phase uses post-hoc explanation methods, such as SHAP, LIME, Morris Sensitivity, and PDP, to explain the cybersickness outcomes. For the global explanation, we use SHAP and Morris sensitivity tool to explain the overall feature ranking based on the overall outcome. On the contrary, the local explanation is based on LIME and PDP for individual feature ranking based on the specific test sample. Once the features are analyzed and ranked using the explanation methods, in the model reduction phase, we use only the top features (1/3 of the features in our case) to retrain the super learner model. This results in a significantly smaller model with fewer hyperparameters, leading to a faster training and inference time. Finally, we deploy the reduced super learner model in an embedded platform for cybersickness detection.

### 3.1 Cybersickness classification using super learner

We implement a super learner-based ensemble model for combining predictions from base models and enhancing the predictions with information from exogenous variables. It is worth mentioning that the super-learning method is based on a general framework of several ensemble algorithms [101]. We start with building base learners and fit the base learners with a meta learner. We use six stacked base learners, tested individually, to achieve the best performance. The idea is to choose a suboptimal classifier to solve the problem, improve the predictive performance, and increase the

**Algorithm 1** Super learning-based cybersickness classification

**Input:** Training dataset: $X_{Train}$,
  Testing dataset: $X_{Test}$ $X_{Test}$,
  List of base learners $B_L$,
  Meta learner $M_L$;
**Output:** $D_L$ = { }
1: **for each** base learners $i \in B_L$ **do**
2:   $F_L = \lambda(X\_train)$
3:   $X\_train' = \theta(X\_train, F_L)$
4: **end for**
5: $F'_L = \lambda(M_L, X\_train')$
6: $P_L = \theta'(F'_L, X\_test)$
7: $D_L = D_L \bigcup P_L$
8: **return** $D_L$

generalization performance of the super learner model. The base learners that we use to build the super learner model are: support vector classifier (SVC) [12], K-nearest neighbors classifier (KNNC)[4], extra trees classifier (ETC)[23], XG boost classifier (XGBC)[11], random forest classifier (RFC)[54], gradient boosting classifier (GBC)[18], and the meta learner classifier is based on a logistic regression classifier [13].

Algorithm 1 shows the overview super learner-based ensemble ML method for cybersickness detection. The algorithm takes several base learners, meta-learners, and training and testing datasets as input and returns the cybersickness level. First, for each of the base learners $B_L$, the training data $X\_train$ is fitted with the learning algorithm function $\lambda$ in a stacking manner to obtain the best-fitted base learners denoted as $F_L$ (Lines 2). Next, the meta-learning training set $X\_train'$ is created based on the prediction function $\theta$ (Line 3). This generates the new dataset for the meta learner using $X\_train$ and $F_L$. The meta learner $M_L$ and meta-learning training set $X\_train'$ are then fitted in the training function $\lambda$ (Line 5). The $\lambda$ function calculates trained value $F_L$ from the ensemble base learner and meta learner. Then, the cybersickness level $D_L$ is predicted from the fitted meta learner $F'_L$ based on the prediction algorithm $\theta$ (Lines 6-7). Finally, the algorithm returns the cybersickness level when this process is complete for all the test data (Line 8).

### 3.2 Cybersickness regression using super learner

We use the same super learner model that we used for classification for the cybersickness regression task, in which base learners we used are support vector regressor (SVR)[12], K- nearest neighbors regressor (KNNR)[4], extra trees regressor (ETR)[23], XG boost regressor (XGBR)[11], random forest regressor (RFR)[54], gradient boosting regressor (GBR)[18]. Then, we used the meta learner regressor, namely the linear regression model, to forecast the cybersickness FMS score in the range of 0 to 10. By comparing the user's current physiological state with the previous physiological state, cybersickness regression regresses the user's ongoing cybersickness FMS score. The cybersickness regression task can formally be defined as follows: Given a history of observed VR data (e.g., eye-tracking data, head-tracking data, physiological signal, gameplay data, etc.) and the FMS score at previous time steps $t - 1$, predict the FMS score at the next time steps $t$. The predicted cybersickness at time $t$ based on the previous time $t - 1$ of physiological signals is denoted by $CSR_t$. For instance, if we predict the cybersickness FMS score at time $t = 20$ seconds, then $CS_t$ can be written as:

$$CSR_t \Rightarrow [P_{t-19}, P_{t-18}, P_{t-17}, P_{t-16} \ldots, P_t]$$

Table 1. Performance of 10-Fold Cross Validation on Cybersickness Severity Classification (non-reduced super learner model)

| Dataset | % Accuracy | % Precision | | | | % Recall | | | | %F1-Score | | | |
|---|---|---|---|---|---|---|---|---|---|---|---|---|---|
| | | None | Low | Medium | High | None | Low | Medium | High | None | Low | Medium | High |
| Gameplay | 82 | 91 | 70 | 56 | 75 | 92 | 90 | 67 | 65 | 91 | 79 | 61 | 71 |
| Bio-physiological | 98 | 95 | - | 96 | 95 | 96 | - | 94 | 97 | 96 | - | 96 | 95 |

Table 2. Performance of 10-Fold Cross Validation on Cybersickness Severity Classification (non-reduced super learner model) for the integrated sensor dataset

| Fusing modalities | % Accuracy | % Precision | | | | % Recall | | | | %F1-Score | | | |
|---|---|---|---|---|---|---|---|---|---|---|---|---|---|
| | | None | Low | Medium | High | None | Low | Medium | High | None | Low | Medium | High |
| Head-tracking | 67 | 71 | 63 | 43 | 64 | 79 | 71 | 48 | 40 | 75 | 67 | 42 | 49 |
| Eye-tracking | 93 | 97 | 91 | 77 | 90 | 96 | 96 | 68 | 81 | 97 | 94 | 73 | 86 |
| Eye + head tracking | 95 | 98 | 95 | 87 | 94 | 99 | 97 | 79 | 90 | 98 | 96 | 83 | 92 |

where $P$ denotes the user's cognitive state.

### 3.3 Cybersickness Explanation

The XAI tools to explain cybersickness outcomes from the super learner model produce visual representations, either as a bar graph or as a set of visualizations either in global or local explanation. We use these graphs to understand model interpretability regarding cybersickness detection and prediction. The explanations can be categorized as global and local explanations. The overall importance ranking (global explanation) of cybersickness detection is visualized as bar graphs. For the local explanation, each sample is randomly chosen from the test dataset, which contains all the features. We use SHapley Additive exPlanations (SHAP) and Morris sensitivity analysis (MSA) to explain the overall feature importance. SHAP assigns feature importance based on a game theoretic approach. On the other hand, Morris sensitivity analysis measures the effect of adjusting one feature at a time, and based on this randomized process; the feature importance is assigned. We are using the results from the global explanation to identify the essential features which help us effectively reduce the feature space.

For the local explanation, we use the Local Interpretable Model-agnostic Explanations (LIME) and partial dependence plots (PDP) tools to explain the individual predictions from the test samples. LIME generates an explanation by approximating the underlying model with an interpretable one to show what feature contributed to the output from that single sample. Similarly, PDP shows the marginal effect one or two features have on the predicted outcome of a machine learning model. This marginal effect can lead to a linear, monotonic, or more complex relation between the output and the feature.

## 4 DATASETS & EXPERMENTAL SETUP

This section explains our experimental setup and datasets to validate our proposed VR-LENSE framework. We used Scikit-Learn [71] for training and evaluating our proposed super learner-based ML model. For explaining the super learner model, we used the SHAP [57] and the InterpretML [66] library. For deploying the super learner-based ML model in a Qualcomm Snapdragon 750G processor-based Android device, we used Android Studio [17], and ONNX [5] library. The super learner-based ML model is trained on an Intel Core i9 Processor and 32GB RAM option with NVIDIA GeForce RTX 3080 Ti GPU.

### 4.1 Datasets

To validate the effectiveness of the proposed VR-LENS framework for cybersickness classification, regression, explanation, and deployment, we used the three datasets, such as integrated sensor [33], bio-physiological [36], and gameplay [74] datasets.

*4.1.1 Integrated sensor dataset.* The integrated sensor dataset [33] contains the eye tracking, head tracking, and physiological signals for 27 participants (Male: 15 and Female: 12) immersed in 5 different VR simulations: Beach City, Road Side, Roller Coaster, SeaVoyage, and Furniture Shop. They recruited a total of 30 participants (Male: 15, Female: 15) to collect the experiment data. However, three participants' data could not be collected due to technical issues (i.e., blacktooth and HTC-Vive wireless adapter black screen issue). Eye tracking, head tracking, and physiological data consist of different subcategories. For instance, in the eye tracking data, the subcategories are Pupil Diameter (left), Pupil Position (x, y, z), Gaze Direction (x, y, z), Convergence Distance, and % of Eye Openness, and for the head tracking data Quaternion Rotation of X, Y, Z, and W axis, respectively. Similarly, for the physiological signals, the subcategories are electrodermal activity (EDA) and HR measurements. This dataset has a total of 20104 samples recorded with a maximum of 7 minutes of VR simulation. In addition, the dataset contains four different cybersickness severity classes: none, low, medium, and high, and the FMS score ranges from 0 to 10, which is used for regression analysis.

*4.1.2 Bio-physiological dataset.* The bio-physiological[36] dataset consists of different physiological signals such as heart rate (HR), breathing rate (BR), heart rate variability (HRV), and galvanic skin response (GSR). The HR, BR, GSR, and HRV data have different subcategories. For instance, in HR data, the subcategories are the percentage of change from resting baseline (PC), minimum inside 3s rolling window (MIN), the maximum value of 3s rolling window (MAX), and moving average of 3s rolling window (AVG). Similarly, for the other bio-physiological signals, the subcategories are the same. They recruited a total of 31 university students from a university class for the experiment (Male: 29, Female:2). Unfortunately, they were unable to collect eight users' data due to some reason (e.g., the battery of the HR sensors died in the middle of the experiment, some users' felt severe sickness during the experiment, etc.,). Therefore, these physiological signals were collected from 23 participants immersed in a virtual roller coaster simulation. This dataset has a total of 24533 samples recorded with a maximum of 897 seconds of VR simulation. The dataset contains three different cybersickness severity classes: *low sickness*, *moderate sickness*, and *acute sickness*. We labeled 'low sickness' as none, 'moderate sickness' as a medium, and 'acute sickness' as a high cybersickness class to reduce ambiguity. The FMS score ranges from 0 to 10 used for regression analysis.

*4.1.3 Gameplay dataset.* The gameplay dataset [74] contains 22 different features from the sources, such as candidate profiles, questionnaires, user field of view, user position, speed of the game in playtime, etc., for 87 participants. This dataset is generated using two VR games, i.e., racing and flight games. However, the data from 35 participants (Male: 26, Female: 9) was collected due to their valid cybersickness. Therefore, the data from participants in the game who answered all virtual reality subjective questionnaires (VRSQ) correctly and completed the whole game interaction The dataset has four cybersickness severity classes are: *none*, *slight*, *moderate*, and *severe*. The FMS score ranges from 0 to 10 and is used for regression analysis. The dataset contains a total of 9391 samples recorded with 5 minutes of VR gameplay simulation [73]. It is worth mentioning that to reduce the ambiguity; we rename these four classes as follows: *none*:none, *slight*: low, *moderate*: medium, and *severe*: high as like integrated sensor dataset.

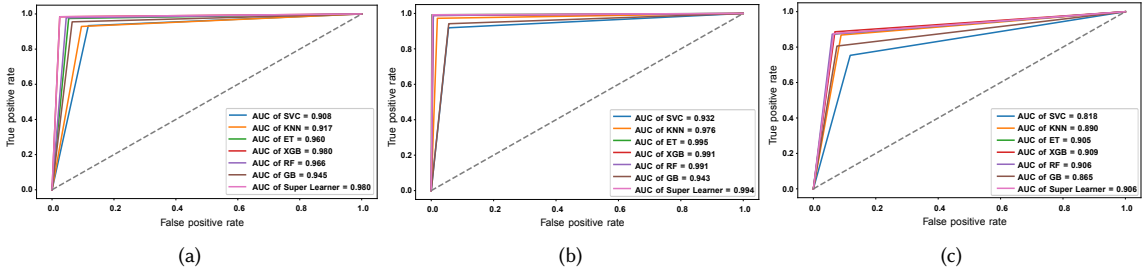

Fig. 3. AUC-ROC curve for Super Learner (**a**) integrated sensor dataset (**b**) gameplay dataset. (**c**) bio-physiological dataset.

### 4.2 Deployment Setup

We used a Samsung A52 5G device running on Android 12 with a Qualcomm Snapdragon 750G processor with up to 2.2 GHz clock speed, 6 GB of RAM, and 128 GB of memory to evaluate our proposed super learner model. Qualcomm Snapdragon processors are typical in VR HMDs, which motivated our choice of this device. In this setup, we simulate the VR setup by injecting integrated sensor data for the model to detect cybersickness severity and measure their outcome and inference time to assess the model's performance.

### 4.3 Performance Metrics

The performance of the super learner-based ensemble model for the cybersickness classification is evaluated using the standard quality metrics such as accuracy, precision, recall, F-1 score, the Area Under the Curve (AUC), and Receiver Operating Characteristic curve (ROC) [1]. Likewise, the performance of the regression models is analyzed using well-known loss functions such as Root Mean Square Error (RMSE), Mean Absolute Error (MAE), Pearson Linear Correlation Coefficient (PLCC) [102], and $R^2$ score. For example, if $y_t$ and $\hat{y}_t$ denote the actual and predicted cybersickness of a candidate at time $t$, respectively, then the RMSE can be defined as follows.

$$RMSE = \sqrt{\frac{1}{|N|} \sum_{y \in S} \sum_{t=1}^{N} (y_t - \hat{y}_t)^2}$$

where $N$ and $S$ represent the total number of samples and time steps, respectively. Mathematically, the MAE can be expressed as follows.

$$MAE = \frac{1}{|N|} \sum_{y \in S} \sum_{t=1}^{N} (y_t - \hat{y}_t)$$

It is worth mentioning that the smaller the MAE and RMSE, the better the regression model. If $\sum(y_t - \hat{y}_t)^2$ and $\sum(y_t - \bar{y})^2$ represent the sum squared regression (SSR) and the total sum of squares (SST) then, the $R^2$ score can be expressed as follows.

$$R^2 = 1 - \frac{\sum(y_t - \hat{y}_t)^2}{\sum(y_t - \bar{y})^2}$$

Consequently, a low $R^2$ value indicates that the regression model does not adequately capture the output variance.

| Dataset | MAE | $R^2$ | RMSE | PLCC |
|---|---|---|---|---|
| **Gameplay** | 0.11 | 0.57 | 0.21 | 0.75 |
| **Bio-physiological** | 0.01 | 0.99 | 0.02 | 0.99 |

(a) For bio-physiological and gameplay datasets

| Fusing modalities | MAE | $R^2$ | RMSE | PLCC |
|---|---|---|---|---|
| **Head-tracking** | 0.72 | 0.49 | 1.08 | 0.70 |
| **Eye-tracking** | 0.23 | 0.92 | 0.43 | 0.96 |
| **Eye + head tracking** | 0.02 | 0.92 | 0.04 | 0.96 |

(b) Integrated sensor dataset

Table 3. Cybersickness regression using non-reduced super learner model (with all features)

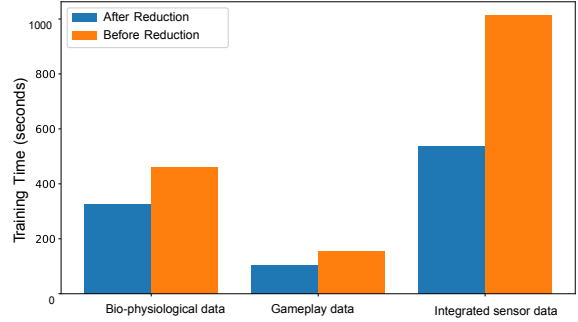

Fig. 4. Training time for Super learned model: non-reduced model (all features) vs. reduced model (reduced features).

## 5 RESULTS

This section presents the results of cybersickness detection, explanation, reduction and deployment.

### 5.1 Cybersickness classification and regression performance with all features

Before applying the XAI-based model reduction, we present the non-reduced super learner model development results with their important statistics, such as training and inference time, cybersickness classification, and regression accuracy. This will help us compare our approach's effectiveness (XAI-based reduction) in the following sections of the paper.

Table 4. Hyperparameters for the super learner (non-reudced vs reduced) model

| Model | | Hyperparameters |
|---|---|---|
| **KNN** | Original | Default |
| | Reduced | no. of neighbors = 2 |
| **SVC** | Original | Default |
| | Reduced | kernel= polynomial |
| **ET** | Original | Default |
| | Reduced | no. of estimators = 50 |
| **RF** | Original | Default |
| | Reduced | no. of estimators = 10 |
| **XGB** | Original | Default |
| | Reduced | no. of estimators=30 |
| **GB** | Original | Default |
| | Reduced | learning rate= 0.05 |

*5.1.1 Cybersickness Detection Model Development.* Table 4 shows the hyperparameters list for the super learner-based cybersickness detection model. The model is trained based on the default hyperparameters from Scikit-learn [72] in which every learner is taken with default hyperparameters values. We use a 10 fold cross-validation technique to train and test the performance of the super learner model similar to [6] in which the dataset is partitioned into $k$ groups (i.e., in our case $k = 10$). Only one partition out of $k$ is utilized for testing the model, while the remaining partitions are used for training. The method is repeated $k$ times, each time picking a new test partition and the remaining $(k − 1)$ partition as a training dataset to eliminate bias. Note we do not use a leave-participant-out validation technique to train and test the performance of our proposed super learner model. The reason is that our dataset is quite large, and leave-participant-out validation is more appropriate for a small dataset since it uses more training samples in each iteration [58]. The training and inference times for integrated sensor, gameplay, and physiological datasets are shown in Figure 4. Training the integrated sensor, gameplay, and physiological datasets using super learner (with all the features) requires 1012, 460, and 154 seconds, respectively.

*5.1.2 Cybersickness Classification.* We summarize the accuracy, precision, recall, and the F-1 score of cybersickness severity classification using the super learner model in Tables 1 and 2. For the integrated sensors, bio-physiological, and gameplay datasets, our super learner model achieves 95%, 98%, and 82% accuracy, respectively. From Table 1, we can observe that the overall performance of the super learner model for the bio-physiological dataset is significantly better than the gameplay dataset in terms of precision, recall, and F1-score for cybersickness severity classification. For example, the physiological dataset obtains a precision value of 95%, 96%, and 95% for the none, medium, and high cybersickness classes. On the other hand, the gameplay dataset obtains the precision value of 91%, 70%, 56%, and 75% for the none, low, medium, and high cybersickness classes. As mentioned in section 4.1.2, the bio-physiological dataset contains only 3 cybersickness classes, namely none, medium, and high. Therefore, there are no results obtained for the low cybersickness class. In addition, Figure 3 presents the AUC-ROC curves for the integrated sensor, physiological, and gameplay datasets using the super-learner model. From Figures 3b and 3c, we observe that the bio-physiological dataset possesses a higher AUC score of 0.994; however, the gameplay dataset has a comparatively lower AUC score of 0.906. In addition, the proposed super-learner model achieves a higher AUC score for all of the datasets in comparison to other baseline models. For instance, the super-learner has an AUC score of 0.994 whereas SVC has an AUC score of 0.932 for the bio-physiological dataset. It is observed that the super-learner performs better than other baseline models except for the XGB classifier model for both bio-physiological and gameplay datasets. This is due to the fact that ensemble methods approximate complex functional relationships of data by combining a set of individual learning algorithms using a meta-learning algorithm [81]. This provides depth insight into the feature and hence, leads to high classification accuracy. Likewise, the accuracy of our proposed super learner model outperforms the DL-based convolutional LSTM model for the bio-physiological dataset [31]. Even though we obtain a great accuracy for the gameplay dataset, but we are unable to compare with symbolic ML-based methods [73]. The reason is that they performed binary classification while we classify the severity of the cybersickness (multiclass classification).

To demonstrate the performance of our proposed model, and compare our work to the state-of-the-art DL model-based cybersickness detection in [33], we use three fusing modalities (*head tracking*, *eye tracking*, and *head + eye tracking*) as shown in Table 2 from the integrated sensors dataset. The proposed super learner model achieves an accuracy of 67%, 93%, and 95% for *head tracking*, *eye tracking*, and *head + eye tracking* fusing modalities, respectively. This outperforms the previously developed deep fusion model proposed in [33] both in terms of *eye tracking* and *head + eye tracking* fusing modalities with accuracies of 80.7% and 87.7%. Likewise, Figure 3a shows the AUC-ROC curves for the integrated sensor datasets using only the *head + eye tracking* fusing modality due to its good accuracy. It is observed that the integrated sensor obtains an AUC score of 0.980, which is higher than other single classifiers except for the XGB classifier model. For instance, the super-learner possesses a higher AUC score of 0.980; however, the baseline model SVC has a comparatively lower AUC score of 0.908.

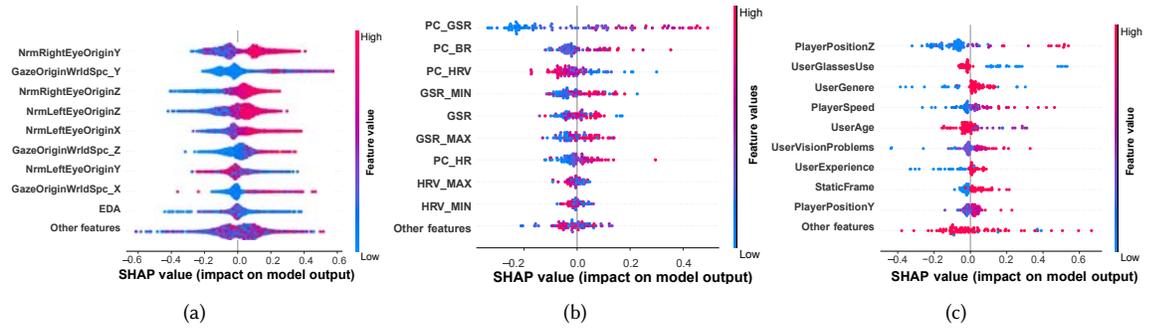

Fig. 5. Overall feature importance using global explanation using SHAP for (**a**) integrated sensor dataset (**b**) bio-physiological dataset (**c**) gameplay dataset.

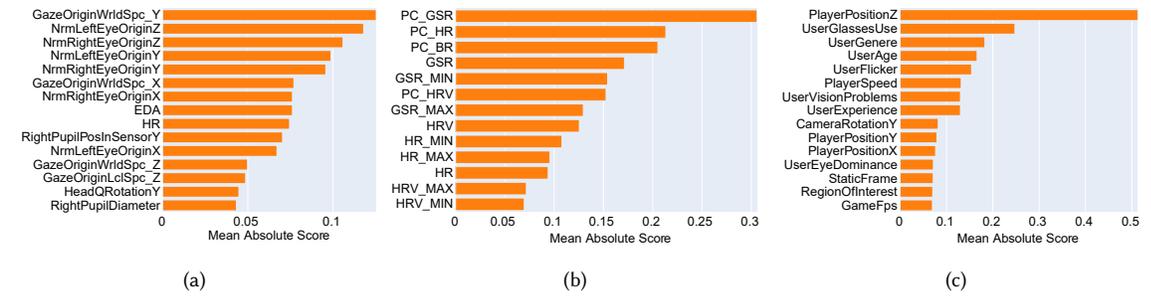

Fig. 6. Overall feature importance using for MSA-based global explanation for: (**a**) integrated sensor dataset (**b**) bio-physiological dataset (**c**) gameplay dataset.

*5.1.3 Cybersickness Regression.* Table 3a and 3b summarizes the MAE, $R^2$, RMSE, and PLCC values for the cybersickness regression using the super learner model on the integrated sensors, bio-physiological, and gameplay datasets. The MAE, $R^2$, RMSE, and PLCC values for the bio-physiological and gameplay datasets are 0.01,0.99,0.02, 0.99 and 0.11, 0.57, 021, 0.75, respectively. Similar to the classification, the regression also uses three fusing modalities (*head tracking, eye tracking, and head + eye tracking*) as shown in Table 3b for the integrated sensors dataset.

The proposed super learner model outperforms the previously reported results [33] in cybersickness regression. For instance, the deep fusion model regression results showed a $R^2$ score value of 0.18, 0.56, and 0.67, while the results from the proposed super learner model show a $R^2$ score value of 0.49, 0.92, and 0.92 for head tracking, eye tracking, and head + eye tracking fusing modalities, respectively. It is worth mentioning that the high $R^2$ value indicates that the regression model performs well in regressing the ongoing cybersickness. The reason behind the good performance of the super learner model is the fact that the super learner model is built on the ensemble technique, which significantly improves the model performance.

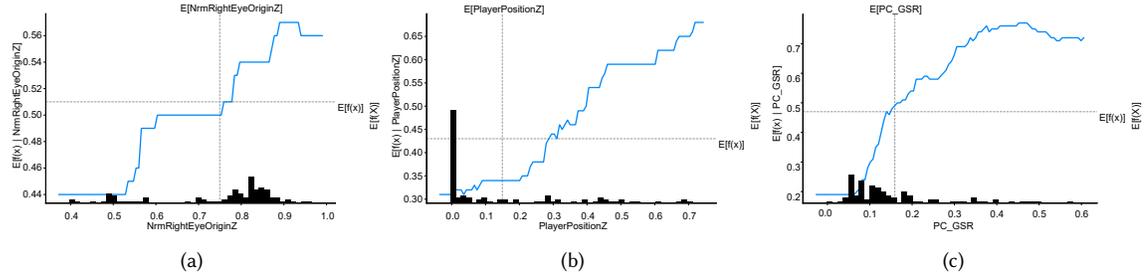

Fig. 7. PDP-based local explanation of cybersickness classification for: (**a**) Integrated sensors dataset. (**b**) Gameplay dataset. (**c**) Bio-physiological dataset.

### 5.2 Model reduction using XAI-based cybersickness explanation

In this section, we apply the post-hoc explanation methods, namely SHAP, Morris sensitivity analysis, LIME, and PDP, to explain the super learner model outcome. These explanations, specifically the global explanation from the SHAP and MSA, are then used to identify the dominating features to reduce the super learner-based cybersickness detection model.

*5.2.1 Cybersickness Global Explanation.* The overall feature importance for cybersickness severity classification using the super learner model with all features for the bio-physiological, gameplay, and integrated sensors datasets are presented in Figure 5 based on SHAP explanation. A shapely value calculates the ranking of the most important features contributing to the cybersickness severity classification, with important features at the top and the least important ones at the bottom. From Figure 5a, we observe that features such as NrmRightEyeOriginY, corresponding to the normalized right eye origin in the Y axis measurement, *GazeOriginWrldSpc_Y*, corresponding to the gaze origin in the world space in the Y axis, and *NrmRightEyeOriginZ,* corresponding to the normalized right eye origin in the Z axis, etc., are the most dominant features in cybersickness severity classification for the integrated sensor dataset. It is worth mentioning that the eye-tracking features have a much stronger influence than the head-tracking features on the cybersickness severity classification. Because eye tracking features contain insightful information such as the type of blink of the user, gaze behavior, and the position of the pupil to track the user's activity [33, 34, 37]. Similarly, from Figure 5 b, it is observed that features such as *PC_GSR* corresponding to the percentage of galvanic skin responses (GSR) measurement, *PC_BR* corresponding to the percentage change of breathing rate (BR) measurement and *PC_HRV* corresponding to the percentage change of heart rate variability (HRV) measurement, etc., have a much stronger influence in the cybersickness classification for the bio-physiological dataset. Likewise, it is observed that the features such as *Player Position Z*, *user glasses use*, and *user gender*, etc., are the most important features in cybersickness classification for the gameplay dataset (Figure 5c). Figure 6 presents the overall feature importance in the cybersickness classification using the Morris sensitivity analysis for the integrated sensor, bio-physiological, and gameplay datasets. Mean absolute score (MAS) is used to calculate the ranking of the most important features contributing to the cybersickness classification. From Figure 6a, we observe that for the integrated sensor dataset, most of the features contributing to cybersickness classification are again eye-tracking features, i.e., GazeOriginWrldSpc_Y, corresponding to the gaze origin in the world space in the Y axis measurement NrmLeftEyeOriginZ, corresponding to the normalized left eye origin in the Z axis measurement NrmRightEyeOriginZ, corresponding to the normalized right eye origin in the Z axis measurement, etc., Similarly, From Figure 6b and c, it is observed that the most predictive features of cybersickness severity classification for the bio-physiological dataset are *PC_GSR*, PC_HR, *PC_BR*, etc., and *player Position Z, user glasses use, user genere*, etc., for the gameplay dataset, respectively.

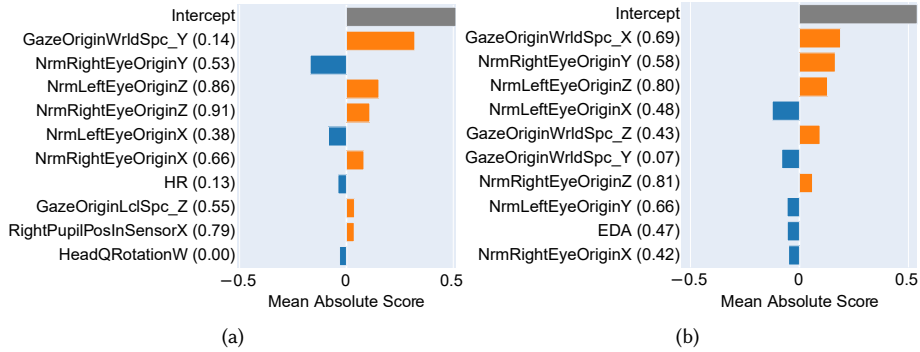

Fig. 8. LIME-based local explanation of cybersickness classification for the integrated dataset (a) explanation for high cybersickness severity, (b) explanation for low cybersickness severity.

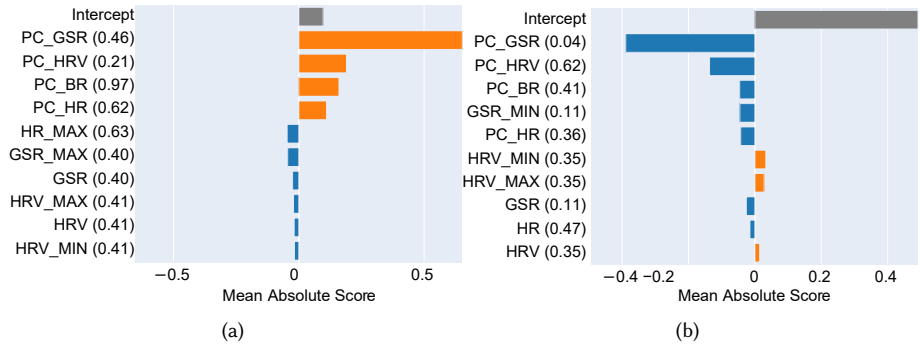

Fig. 9. LIME-based local explanation of cybersickness classification for the bio-physiological dataset (a) explanation for high cybersickness severity, (b) explanation for low cybersickness severity.

5.2.2 *Cybersickness Local Explanation.* The results of the local explanation utilizing LIME for the integrated sensor dataset are shown in Figure 8. Figure 8a shows the high cybersickness severity classification, in which the yellow and black colored bars denote cybersickness probabilities MAS for that individual outcome. The $x$-axis represents the model's output MAS value are log odds (the probabilities of feature importance in prediction), and the $y$-axis lists the model's features. Most features contribute to the negative impact indicated as yellow bars; hence, an accurate decision is made for the high cybersickness severity class. It is observed that the eye tracking feature *GazeOriginWrldSpc_Y*, corresponding to the gaze origin from world space in the Y axis measurement, is the most influential feature for high cybersickness severity classification, which has the highest MAS value, nearly 0.4. For example, in Figure 8b, none cybersickness severity classification has the negative MAS value for most of the features, which indicates that most of the features contribute to the positive impact. Most of the features except *EDA* corresponding to electrodermal activity measurement belong to eye tracking features; thus, an appropriate decision is established for none cybersickness severity classification. Similarly, the local explanation of the classified cybersickness for the bio-physiological dataset is shown in Figure 9. In Figure 9a and Figure 9b, we observe that most features contribute to a high cybersickness severity class corresponding to the features *PC_GSR* corresponding to the percentage of GSR measurement, *PC_HRV* corresponding to the percentage of HRV measurement, etc., and for none cybersickness severity class, the dominating features are, i.e., *HRV_MIN* corresponding to the minimum HRV measurement, *HRV_MAX* corresponding to the maximum HRV measurement, etc., Likewise, the local explanation of the classified cybersickness for the gameplay dataset is shown in Figure 10. From Figure 10a, it is observed that the most influential features for the high cybersickness severity class belong to eye-tracking features such as *user genere*, *user age*, etc. Consequently, Figure 10b shows that most of the features that influence the positive outcome (none cybersickness severity class) are player position Z, user age, static frame, *user vision problems*, etc., The cybersickness classification for the gameplay dataset has low accuracy, as discussed in Section 5.1.2. So, there is a wrong explanation of the features such as *user age*, *player position Y* in both positive and negative outcomes. Such local explanation of features provides insights into the classification/misclassification results and thus builds trust in the model outcome to make appropriate decisions.

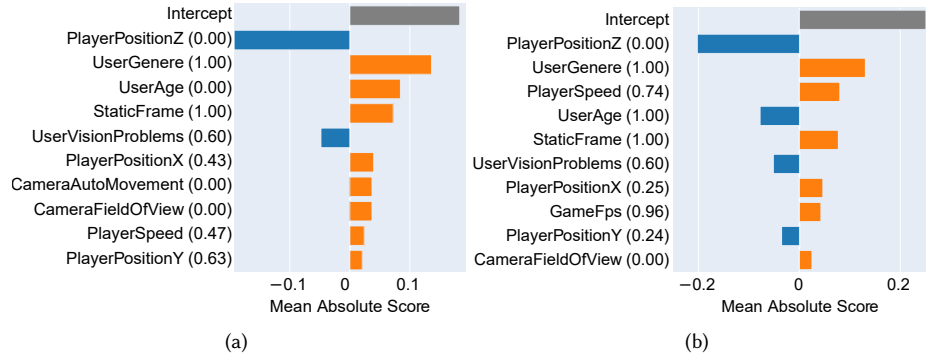

Fig. 10. LIME based-local explanation of cybersickness classification for the gameplay dataset (a) explanation for high cybersickness severity, (b) explanation for low cybersickness severity.

We analyze the relation between top features and the model output utilizing a PDP-based local explanation to provide a deeper insight into the cybersickness severity classification explanation. Figure 7 presents the PDP results of the top feature for the integrated sensor, bio-physiological, and gameplay datasets. PDP explanation aims to identify the partial relationship between a set of given features and the corresponding predicted value. Figure 7a shows that the eye tracking feature *NrmRightEyeOriginZ*, corresponding to the normalized right eye origin in the Z axis measurement, has a positive relationship with the cybersickness classification (the higher the value, the higher the probability of cybersickness). Similarly, Figure 7b and c shows the feature *player position on the z axis* for the gameplay and *PC_GSR* for the bio-physiological dataset have a positive relationship with the cybersickness classification.

This section presents feature selection, cybersickness model reduction, and deployment in an embedded platform.

*5.2.3 Model reduction for deployment.* Using the MSA and SHAP-based (global) explanation results described in the previous section, we first identify the top 1/3 of the features and retrain the super learner-based ensemble model. We obtained the ratio of 1/3 by the trial and error method for our super learner model and datasets. For instance, we took the top 15 features out of 43 from the integrated sensors dataset, the top 4 features out of 13 features from the bio-physiological dataset, and the top 10 features out of 20 features from the gameplay datasets. Table 6 shows the selected features through SHAP and MSA-based global explanation for retraining the proposed super-learner-based ensemble model. Likewise, Table 4 shows the list of hyperparameters for the retrained super learner model.

Table 5. Inference time from Super learner deployment: beofre reduction (all features) vs. after reduction (reduced features)

| Sample size | Inference Time in seconds | |
|---|---|---|
| | Before feature reduction | After feature reduction |
| 1 | 0.15 | 0.07 |
| 10 | 1.2 | 0.72 |
| 100 | 11.8 | 7.6 |
| 500 | 54.7 | 38.04 |

Table 6. List of selected features for retraining the super learner model through global explanation

| Dataset | Selected Features from SHAP and MSA -based Global Explanation |
|---|---|
| Integrated sensor | NrmRightEyeOriginZ; NrmLeftEyeOriginY; NrmRightEyeOriginX; NrmRightEyeOriginY; NrmLeftEyeOriginZ; NrmLeftEyeOriginX; NrmSRLeftEyeGazeDirX; NrmSRLeftEyeGazeDirY; NrmSRRightEyeGazeDirY; GazeOriginWrldSpc_Y; GazeOriginWrldSpc_Z; RightPupilDiameter; HeadQRotationW; HeadQRotationY; HeadEulX |
| Gameplay | PlayerPositionZ; UserGlassesUse; UserGenere; UserAge; PlayerSpeed; UserVisionProblems; UserExperience; PlayerPositionX; StaticFrame; CameraRotationY |
| Bio-physiological | PC_GSR; PC_BR; GSR_MIN; PC_HRV |

Table 7. Performance of 10-Fold Cross Validation on Cybersickness Severity Classification using the reduced super learner model (reduced features)

| Dataset | Feature Count | | % Acc. | % Precision | | | | % Recall | | | | %F1 Score | | | |
|---|---|---|---|---|---|---|---|---|---|---|---|---|---|---|---|
| | Original | Reduced | | None | Low | Medium | High | None | Low | Medium | High | None | Low | Medium | High |
| Gameplay | 22 | 10 | 81 | 88 | 81 | 79 | 94 | 95 | 89 | 75 | 80 | 91 | 79 | 78 | 80 |
| Integrated Sensors | 43 | 15 | 96 | 99 | 95 | 91 | 95 | 98 | 98 | 97 | 95 | 99 | 96 | 89 | 94 |
| Bio-Physiological | 14 | 4 | 98 | 99 | - | 97 | 99 | 99 | - | 98 | 96 | 99 | - | 98 | 99 |

*5.2.4 Performance of reduced Super learner model.* The total size of the deployed model in the Samsung A52 5G device is 133, 609 KB for the non-reduced and 90, 917 KB for the reduced super learner models. Figure 4 shows the improvement in training time using the reduced super learner model for the reduced bio-physiological, gameplay, and integrated sensor datasets. It is observed that the training time is improved significantly for the reduced super learner model by 1.53X, 1.46X, and 1.91X for the bio-physiological, gameplay, and integrated sensors datasets, respectively. The deployed reduced and non-reduced super learner models' inference times are shown in Table 5. For instance, inference on the deployed non-reduced super learner model requires 0.15, 1.2, 11.8, and 54.7 seconds for the 1, 10, 100, and 500 samples, respectively. On the other hand, the reduced super learner model requires only 0.07, 0.72, 7.6, and 38.04 for the 1, 10, 100, and 500 samples which are 2.15X, 1.67X, 1.58X, and 1.44X faster than the non-reduced super learner model.

Table 8. Cybersickness regression using reduced super learner model (with reduced features) for bio-physiological, integrated sensor, and gameplay datasets

| Dataset | MAE | $R^2$ | RMSE | PLCC |
|---|---|---|---|---|
| **Bio-physiological** | 0.03 | 0.95 | 0.06 | 0.97 |
| **Integrated Sensors** | 0.02 | 0.95 | 0.03 | 0.97 |
| **Gameplay** | 0.12 | 0.49 | 0.22 | 0.70 |

*5.2.5 Cybersickness classification for the reduced super learner model.* As mentioned earlier, we deploy the super learner model in embedded hardware for classifying the integrated sensor dataset. Then, we simulate the super learner model for the rest of the dataset to evaluate their performance.

Table 7 summarizes the accuracy, precision, recall, and F-1 scores of cybersickness classification using the reduced order super learner model for integrated sensors, bio-physiological, and gameplay datasets. For instance, cybersickness classification for the integrated sensor dataset using the reduced super learner model exhibits 96% accuracy. In addition, the cybersickness classification accuracy for the bio-physiological dataset is 98%, which is also slightly higher than the accuracy of the non-reduced super learner model for the bio-physiological dataset (see Table 1). However, the cybersickness classification accuracy of the reduced super learner model for the gameplay dataset slightly decreased by 1.2% compared to their non-reduced version. Furthermore, other performance metrics, such as precision, recall, and F1-score for the none, low, medium, and high cybersickness classes, slightly increased for the reduced super learner model compared to their non-reduced version for the three datasets. For instance, the precision score for the none, low, medium, and high cybersickness classes for the reduced super learner model using integrated sensor dataset are 99%, 95%, 91%, and 95%, which is slightly better than the non-reduced super learner model. Likewise, the recall score for the none, medium, and high cybersickness classes for the reduced super learner model are 99%, 98%, 97%, and 95%, in which medium class 1.23 higher than the non-reduced model. In addition, for the other datasets, specifically for the bio-physiological dataset, the reduced model improved the precision, recall, and F1 score for all classes.

*5.2.6 Cybersickness regression for the reduced super learner model.* Table 8 shows the performance of cybersickness regression using the reduced order super learner model for the integrated sensors, bio-physiological, and gameplay datasets. For instance, the MAE, ($R^2$), RMSE, and PLCC values for the reduced super learner model with integrated sensor dataset are 0.02, 0.95, 0.04, and 0.97 and 0.03, 0.95, 0.06, and 0.97 for the bio-physiological dataset, respectively. This reduced super learner model for integrated and bio-physiological datasets has significantly improved the MAE, ($R^2$), RMSE, and PLCC values. On the other hand, it is observed that for the gameplay dataset MAE, ($R^2$), RMSE, and PLCC values are 0.12, 0.49, 0.22, and 0.70, which is slightly better than the non-reduced model. This is because the gameplay data contains the features, mostly categorical features. .

## 6 DISCUSSION

This section briefly discusses the results obtained using the VR-LENS framework. The SHAP and MSA-based global explanation reveal that for the integrated sensor dataset, features such as *normal eye origin*, *gaze origin*, *pupil diameter*, etc., are the most influential features for causing cybersickness. Similarly, for the bio-physiological and gameplay datasets features such as *PC_GSR*, *PC_BR* and *PC_HRV*, etc., and *player Position Z*, *user glasses use*, *user genere*, etc., respectively. On the contrary, the LIME and PDP-based local explanations of specific predictions offered useful insight for each sample. Consequently, we ranked the features' importance from the global explanation using SHAP and MSA, and important features were used to retrain the super learner model. Our results suggest that the SHAP and MSA-guided reduced super learner model result in significantly faster training times. Furthermore, the deployed reduced super learner model in a Qualcomm Snapdragon 750G processor-based Samsung A52 5G device shows faster inference time for real-time cybersickness detection without sacrificing accuracy. For instance, the deployed reduced super learner cybersickness model, which was trained with only 1/3 of the features compared to its non-reduced version, classified the cybersickness severity with an accuracy of 96%. Similarly, while regressing to cybersickness, the reduced super learner obtained an RMSE value of 0.03, which is 25.5% less than its non-reduced version for the integrated sensor dataset. It is worth mentioning that the reduced super learner performed well for the bio-physiological dataset in both cybersickness classification and regression. In contrast, the gameplay dataset performed poorly in both cybersickness classification and regression. The reason is that the gameplay dataset contains features from mostly users' profile data, which doesn't provide any useful insight into cybersickness. However, the integrated sensor data, such as eye-blink rate, pupil diameter, HR, etc., provided more insights into user behavior regarding cybersickness.

The accuracy of our proposed super learner model outperforms several state-of-the-art works in ML and DL-based cybersickness detection. For instance, Islam et al. [34] used a deep temporal convolutional network (DeepTCN) to forecast the cybersickness FMS score (on a scale from 0–10) with an RMSE value of 0.49, based on eye tracking, heart rate, and galvanic skin response data. In contrast, Dennison et al. [14] reported accuracy of 78% and $R^2$ values 75% using the bio-physiological data. Our super learner model's classification and regression accuracy outperform these works. There also exist other works which are relevant to our work. For instance, Qu et al. [75], Kim et al. [46], Garcia-Agundez et al. [20], and Jeong et al. [39] reported cybersickness detection accuracy of 96.85%, , 89.16%, 82% and 94.02%, respectively, using bio-physiological and EEG/ECG signals.

Even though there are several works in cybersickness detection methods, to date, only a few studies have been conducted on identifying the causes of cybersickness [31, 33, 46, 67]. However, to the best of our knowledge, no prior work exists on applying XAI to explain the cybersickness from black-box ML models, reducing ML model size using XAI, and deploying them on embedded devices. Indeed, XAI-based explanations can help researchers understand the reasons behind correct and incorrect cybersickness classification and can be further utilized to develop effective cybersickness reduction methods. Therefore, we believe that the proposed XAI-based cybersickness model reduction and deployed model can help researchers to automate the cybersickness detection in real-time on standalone VR headsets and improve the usability of the VR.

## 7 LIMITATIONS AND FUTURE WORKS

Although our proposed XAI-based super learner model for cybersickness detection, feature reduction, and deployment method outperformed the state-of-the-art cybersickness detection models, our approach has a few limitations. For instance, we demonstrated the effectiveness of our proposed XAI-based model reduction method with a fast training and inference time and also deployed the model in a Qualcomm Snapdragon processor-based (state-of-the-art VR HMDs use Qualcomm Snapdragon processors) Samsung A52 device. However, we did not deploy our models on an actual VR headset. Therefore, it is hard to explain what type of sensors would perform well in cybersickness detection. However, based on our proposed XAI-based method, eye-tracking sensors are much more efficient than a head-tracking sensors for cybersickness prediction. Consequently, it is worth mentioning that external sensors (e.g., heart rate, galvanic skin response, electroencephalogram) can limit VR locomotion and 3D-object manipulation during the immersion. In addition, these sensors often require tethering and affixing to the users' hands.

Furthermore, cybersickness might affect different people based on their unique characteristics, VR environment, and gender [86]. For instance, Females whose Interpupillary distance (IPD) could not be properly fit into the VR headset often suffered from high cybersickness and did not fully recover within a short time [28, 86]. Therefore, in the future, we plan to conduct further research with people from broader demographic backgrounds and of equal gender representation. Also, this work uses only eye-tracking, head-tracking, bio-physiological signals, and users' profile data to detect cybersickness. In the future, we plan to investigate the effect of stereo images and stereoscopic video data to detect cybersickness with explainability and deploy it in a realistic VR headset.

## 8 CONCLUSION

In this work, we proposed the VR-LENS framework, an XAI-based framework for cybersickness detection through a super learner-based ensemble ML model with explanations and deployment in embedded devices. Specifically, we developed a super learner-based ensemble ML model for cybersickness detection. Then applied, XAI to explain the cybersickness and reduce the feature space and model size for deploying in the embedded device. We illustrated the effectiveness of our proposed method using three datasets, i.e., the integrated sensor, bio-physiological, and gameplay datasets. Our global explanation results revealed that eye-tracking features are the most influential for causing cybersickness in the integrated sensor dataset. Consequently, for the bio-physiological dataset, the GSR, HR, and for the gameplay dataset, the player Position and user glasses of the user are the most influential feature in causing cybersickness. Furthermore, we identified more helpful insight for each sample (misclassification instances) using the local explanation. Finally, based on the XAI-based feature ranking, we significantly reduced the super learner model size and deployed it on a Qualcomm Snapdragon processor-based Samsung A52 device system. The deployed super learner model significantly reduced the training time (up to 1.91X) and inference time (up to 2.46X). For instance, our deployed reduced super learner model could classify and regress the cybersickness with an accuracy of 96% and RMSE of 0.03 for the integrated sensor dataset, which outperforms the state-of-art works. To our knowledge, this is the first work applying XAI to explain a super learner-based ensemble model, reduce the model size, and deploy it in an embedded device. We believe this research will be helpful for future researchers working on cybersickness detection, mitigation, and real-time prediction of cybersickness in standalone VR headsets.

## ACKNOWLEDGMENTS

This material is based upon work supported by the National Science Foundation under Award Numbers: CNS-2114035. Any opinions, findings, conclusions, or recommendations expressed in this publication are those of the authors and do not necessarily reflect the views of the National Science Foundation.

IUI '23, March 27–31, 2023, Sydney, Australia    Kundu and Hoque et al.
[35] Rifatul Islam, Yonggun Lee, Mehrad Jaloli, Imtiaz Muhammad, Dakai Zhu, Paul Rad, Yufei Huang, and John Quarles. 2020. Automatic detection and prediction of cybersickness severity using deep neural networks from user's physiological signals. In *2020 IEEE International Symposium on Mixed and Augmented Reality (ISMAR)*. IEEE, 400–411.

[36] Rifatul Islam, Yonggun Lee, Mehrad Jaloli, Imtiaz Muhammad, Dakai Zhu, Paul Rad, Yufei Huang, and John Quarles. 2020. Automatic Detection and Prediction of Cybersickness Severity using Deep Neural Networks from user's Physiological Signals. In *2020 IEEE International Symposium on Mixed and Augmented Reality (ISMAR)*. 400–411. https://doi.org/10.1109/ISMAR50242.2020.00066

[37] Dayoung Jeong and Kyungsik Han. 2022. Leveraging multimodal sensory information in cybersickness prediction. In *Proceedings of the 28th ACM Symposium on Virtual Reality Software and Technology*. 1–2.

[38] Daekyo Jeong, Sangbong Yoo, and Jang Yun. 2019. Cybersickness analysis with eeg using deep learning algorithms. In *2019 IEEE Conference on Virtual Reality and 3D User Interfaces (VR)*. IEEE, 827–835.

[39] D. Jeong, S. Yoo, and J. Yun. 2019. Cybersickness Analysis with EEG Using Deep Learning Algorithms. In *2019 IEEE Conference on Virtual Reality and 3D User Interfaces (VR)*. 827–835. https://doi.org/10.1109/VR.2019.8798334

[40] Weikuan Jia, Meili Sun, Jian Lian, and Sujuan Hou. 2022. Feature dimensionality reduction: a review. *Complex & Intelligent Systems* (2022), 1–31.

[41] Weina Jin, Jianyu Fan, Diane Gromala, and Philippe Pasquier. 2018. Automatic prediction of cybersickness for virtual reality games. In *2018 IEEE Games, Entertainment, Media Conference (GEM)*. IEEE, 1–9.

[42] Robert S Kennedy, Norman E Lane, Kevin S Berbaum, and Michael G Lilienthal. 1993. Simulator sickness questionnaire: An enhanced method for quantifying simulator sickness. *The international journal of aviation psychology* 3, 3 (1993), 203–220.

[43] Behrang Keshavarz and Heiko Hecht. 2011. Validating an efficient method to quantify motion sickness. *Human factors* 53, 4 (2011), 415–426.

[44] Behrang Keshavarz, Katlyn Peck, Sia Rezaei, and Babak Taati. 2022. Detecting and predicting visually induced motion sickness with physiological measures in combination with machine learning techniques. *International Journal of Psychophysiology* (2022).

[45] Hyun K Kim, Jaehyun Park, Yeongcheol Choi, and Mungyeong Choe. 2018. Virtual reality sickness questionnaire (VRSQ): Motion sickness measurement index in a virtual reality environment. *Applied ergonomics* 69 (2018), 66–73.

[46] Jinwoo Kim, Woojae Kim, Heeseok Oh, Seongmin Lee, and Sanghoon Lee. 2019. A deep cybersickness predictor based on brain signal analysis for virtual reality contents. In *Proceedings of the IEEE/CVF International Conference on Computer Vision*. 10580–10589.

[47] Young Youn Kim, Eun Nam Kim, Min Jae Park, Kwang Suk Park, Hee Dong Ko, and Hyun Taek Kim. 2008. The application of biosignal feedback for reducing cybersickness from exposure to a virtual environment. *Presence: Teleoperators and Virtual Environments* 17, 1 (2008), 1–16.

[48] R Kottaimalai, M Pallikonda Rajasekaran, V Selvam, and B Kannapiran. 2013. EEG signal classification using principal component analysis with neural network in brain computer interface applications. In *2013 IEEE international conference on emerging trends in computing, communication and nanotechnology (ICECCN)*. IEEE, 227–231.

[49] Ripan Kumar Kundu, Rifatul Islam, Prasad Calyam, and Khaza Anuarul Hoque. 2022. TruVR: Trustworthy Cybersickness Detection using Explainable Machine Learning. *arXiv preprint arXiv:2209.05257* (2022).

[50] Ripan Kumar Kundu, Akhlaqur Rahman, and Shuva Paul. 2021. A Study on Sensor System Latency in VR Motion Sickness. *Journal of Sensor and Actuator Networks* 10, 3 (2021), 53.

[51] Joseph J LaViola Jr. 2000. A discussion of cybersickness in virtual environments. *ACM Sigchi Bulletin* 32, 1 (2000), 47–56.

[52] Tae Min Lee, Jong-Chul Yoon, and In-Kwon Lee. 2019. Motion sickness prediction in stereoscopic videos using 3d convolutional neural networks. *IEEE transactions on visualization and computer graphics* 25, 5 (2019), 1919–1927.

[53] Ajey Lele. 2013. Virtual reality and its military utility. *Journal of Ambient Intelligence and Humanized Computing* 4, 1 (2013), 17–26.

[54] Andy Liaw and Matthew Wiener. 2002. Classification and Regression by randomForest. *R News* 2, 3 (2002), 18–22. https://CRAN.R-project.org/doc/Rnews/

[55] Chin-Teng Lin, Shu-Fang Tsai, and Li-Wei Ko. 2013. EEG-based learning system for online motion sickness level estimation in a dynamic vehicle environment. *IEEE transactions on neural networks and learning systems* 24, 10 (2013), 1689–1700.

[56] Yi-Tien Lin, Yu-Yi Chien, Hsiao-Han Wang, Fang-Cheng Lin, and Yi-Pai Huang. 2018. 65-3: The Quantization of Cybersickness Level Using EEG and ECG for Virtual Reality Head-Mounted Display. In *SID Symposium Digest of Technical Papers*, Vol. 49. Wiley Online Library, 862–865.

[57] Scott M Lundberg and Su-In Lee. 2017. A Unified Approach to Interpreting Model Predictions. In *Advances in Neural Information Processing Systems 30*, I. Guyon, U. V. Luxburg, S. Bengio, H. Wallach, R. Fergus, S. Vishwanathan, and R. Garnett (Eds.). Curran Associates, Inc., 4765–4774. http://papers.nips.cc/paper/7062-a-unified-approach-to-interpreting-model-predictions.pdf

[58] Måns Magnusson, Michael Andersen, Johan Jonasson, and Aki Vehtari. 2019. Bayesian leave-one-out cross-validation for large data. In *International Conference on Machine Learning*. PMLR, 4244–4253.

[59] Sannia Mareta, Joseph Manuel Thenara, Rafael Rivero, and May Tan-Mullins. 2022. A study of the virtual reality cybersickness impacts and improvement strategy towards the overall undergraduate students' virtual learning experience. *Interactive Technology and Smart Education* ahead-of-print (2022).

[60] Nicolas Martin, Nicolas Mathieu, Nico Pallamin, Martin Ragot, and Jean-Marc Diverrez. 2020. Virtual reality sickness detection: An approach based on physiological signals and machine learning. In *2020 IEEE International Symposium on Mixed and Augmented Reality (ISMAR)*. IEEE, 387–399.

[61] Moch Asyroful Mawalid, Alfi Zuhriya Khoirunnisa, Mauridhi Hery Purnomo, and Adhi Dharma Wibawa. 2018. Classification of EEG signal for detecting cybersickness through time domain feature extraction using Naïve bayes. In *2018 International Conference on Computer Engineering, Network and Intelligent Multimedia (CENIM)*. IEEE, 29–34.
22